\documentclass[conference]{IEEEtran}
\IEEEoverridecommandlockouts
\usepackage{cite}
\usepackage{amsmath,amssymb,amsfonts}
\usepackage{algorithmic}
\usepackage{algorithm}
\usepackage{graphicx}
\usepackage{url}
\usepackage{textcomp}
\usepackage[hidelinks]{hyperref}
\def\BibTeX{{\rm B\kern-.05em{\sc i\kern-.025em b}\kern-.08em
    T\kern-.1667em\lower.7ex\hbox{E}\kern-.125emX}}

\makeatletter
\newcommand{\linebreakand}{%
  \end{@IEEEauthorhalign}
  \hfill\mbox{}\par
  \mbox{}\hfill\begin{@IEEEauthorhalign}
}
\makeatother

\begin{document}

\title{Latent Sculpting for Zero-Shot Generalization: A Manifold Learning Approach to Out-of-Distribution Anomaly Detection\\
}

\author{
\IEEEauthorblockN{\textbf{Rajeeb Thapa Chhetri}}
\IEEEauthorblockA{\textit{School of Liberal Arts} \\
\textit{Mercy University}\\
555 Broadway, Dobbs Ferry, NY \\
rthapachhetri@mercy.edu}
\and
\IEEEauthorblockN{\textbf{Saurab Thapa}}
\IEEEauthorblockA{\textit{School of Liberal Arts} \\
\textit{Mercy University}\\
555 Broadway, Dobbs Ferry, NY \\
sthapa4@mercy.edu}
\linebreakand % <--- This forces the exact 2x2 grid break!
\IEEEauthorblockN{\textbf{Avinash Kumar}}
\IEEEauthorblockA{\textit{School of Liberal Arts} \\
\textit{Mercy University}\\
555 Broadway, Dobbs Ferry, NY \\
akumar23@mercy.edu}
\and
\IEEEauthorblockN{\textbf{Zhixiong Chen}}
\IEEEauthorblockA{\textit{School of Liberal Arts} \\
\textit{Mercy University}\\
555 Broadway, Dobbs Ferry, NY \\
zchen@mercy.edu}
}

\maketitle
\thispagestyle{empty}
\pagestyle{empty}

\begin{abstract}
Detecting previously unseen attacks remains a major challenge for machine learning-based intrusion detection systems. Deep models trained on network traffic often achieve high accuracy on known attacks but fail under distributional shift because their decision boundaries are tightly coupled to the training data distribution. We introduce Latent Sculpting, a two-stage anomaly detection framework that improves robustness by explicitly structuring the latent representation before density estimation. The first stage trains a Transformer-based tabular encoder using a novel Binary Latent Sculpting loss, which encourages benign traffic to form a compact latent cluster while enforcing separation from anomalous patterns. The second stage fits a Masked Autoregressive Flow to the resulting latent space to produce calibrated probabilistic anomaly scores. Under a strict zero-shot evaluation protocol on the CIC-IDS-2017 benchmark, Stage 1 attains an F1-score of 0.98 on known attacks, while Stage 2---evaluated at the balanced threshold (85th-percentile)---achieves a zero-shot OOD F1-score of 0.867 and AUROC of 0.913. The model successfully detects difficult distribution shifts including stealthy infiltration attacks (78.7\% recall, peaking at 97.2\%) and low-volume DoS variants (\textgreater{}94\% recall), scenarios where conventional approaches often fail. Our results suggest that explicitly separating latent geometry learning from density modeling provides a stable approach for detecting zero-day cyber threats.
\end{abstract}


\section{Introduction}
Modern Network Intrusion Detection Systems (NIDS) operating on high-dimensional tabular data face a critical vulnerability: the inability to reliably detect zero-day, Out-of-Distribution (OOD) attacks. While machine learning has advanced signature-based detection, the cybersecurity landscape requires models capable of true zero-shot generalization to identify uncatalogued intrusions \cite{sharafaldin2018toward}.

Current methodologies fail to achieve this due to a phenomenon we term ``generalization collapse.'' Supervised deep learning architectures, including recent tabular Transformers \cite{gorishniy2021revisiting}, optimize purely for class separation. They construct precise decision boundaries around known distributions but lack strict topological constraints in their latent spaces. Consequently, novel OOD anomalies seamlessly overlap with benign representations, leading to catastrophic overconfidence and near-total detection failure for zero-day threats \cite{hendrycks2016baseline}. Conversely, purely unsupervised anomaly detection models attempt to map the baseline benign distribution to identify deviations. However, without the structural guidance of labeled data during representation learning, these models struggle to resolve the highly nonlinear, multi-modal nature of tabular network traffic, yielding unmanageable false-positive rates.

We hypothesize that resolving generalization collapse requires explicitly decoupling topological manifold structuring from probabilistic density estimation. To this end, we propose \textbf{Latent Sculpting}, a hierarchical, two-stage representation learning architecture. By enforcing explicit structural boundaries prior to density estimation, our framework provides a mathematically stable substrate for isolating zero-day distributional shifts. 

Our core contributions are:
\begin{itemize}
    \item \textbf{Binary Latent Sculpting Loss:} A novel optimization objective that forces a Transformer-based tabular encoder to aggressively condense benign traffic into a dense, low-entropy hypersphere, while simultaneously enforcing a strict geometric minimum-distance margin for known anomalies.
    \item \textbf{Two-Stage Manifold Density Estimation:} The coupling of our structural encoder with a Masked Autoregressive Flow (MAF) \cite{papamakarios2017masked}. The MAF projects the topologically optimized Stage-1 benign manifold into a tractable probabilistic space, allowing for exact, threshold-based likelihood estimation of novel threats.
    \item \textbf{State-of-the-Art Zero-Shot Performance:} Evaluated on the high-dimensional CIC-IDS-2017 benchmark using a rigorous zero-shot protocol (deliberately withholding complex attack classes during training). Averaged across three random initialization seeds, Latent Sculpting maintains near-perfect accuracy on known signatures ($F1 = 0.980 \pm 0.000$). Crucially, it achieves an OOD zero-shot $F1$-Score of $0.867 \pm 0.021$ and an AUROC of $0.913 \pm 0.010$ at an 85th-percentile threshold.

 Additionally, ablation experiments confirm that both the sculpting loss and density estimation are necessary for optimal performance.
        \item \textbf{Detection of Stealthy Intrusions:} The architecture overcomes historic baseline failures on stealthy shifts, achieving an average recall of $78.7\%$ (peaking at $97.2\%$) on Infiltration attacks and over $94\%$ on low-volume DoS variations.
\end{itemize}

\section{Proposed Methodology}

To address the limitations of standard supervised optimization in high-dimensional domains, we propose a two-stage hierarchical architecture. Our framework explicitly decouples the topological structuring of the latent space from probabilistic density estimation. The pipeline consists of a Tabular Transformer encoder trained via a novel Binary Latent Sculpting Loss, followed by a Masked Autoregressive Flow (MAF) for exact likelihood calculation.

\subsection{Structured Latent Space via Tabular Transformers}
Network intrusion data is inherently tabular, lacking the explicit spatial or sequential relationships found in images or text. To capture complex inter-feature correlations, we implement a Tabular Transformer Encoder.

Let the input vector be $x \in \mathbb{R}^D$, where $D=71$ represents the distinct network features. Standard neural networks project this entire vector simultaneously. Instead, our architecture treats each feature as a discrete token. A feature embedding layer first projects each scalar feature $x_i$ into a dense vector representation $v_i \in \mathbb{R}^{d_{\mathrm{model}}}$ (where $d_{\mathrm{model}} = 64$). 

Because tabular features possess strict semantic identities (e.g., index 0 always represents ``Duration''), we introduce learnable Positional Embeddings $P \in \mathbb{R}^{D \times d_{\mathrm{model}}}$. Recent literature on deep tabular models \cite{gorishniy2021revisiting, somepalli2021saint} demonstrates that injecting positional parameters allows the self-attention mechanism to retain feature identity regardless of input arrangement, effectively acting as a ``feature tokenizer''. The sequence of embedded tokens is passed through multi-head self-attention layers, and a global average pooling operation collapses the sequence into a single, cohesive representation. A final linear head projects this into the latent vector $z \in \mathbb{R}^d$ ($d=64$).

\subsection{Binary Latent Sculpting Loss}
The cornerstone of our architecture is the Binary Latent Sculpting Loss, a novel objective designed to aggressively prevent generalization collapse. Standard Cross-Entropy optimization draws unbounded hyperplanes between classes, leaving vast regions of ``negative space'' where Out-of-Distribution (OOD) data can trigger high-confidence misclassifications. Our loss function restricts the benign distribution to a mathematically compact manifold while establishing a strict geometric exclusion zone for anomalies.

For a mini-batch of $N$ latent vectors $z_i$ and their corresponding binary labels $y_i \in \{0, 1\}$ (where $0$ denotes benign and $1$ denotes anomalous), we dynamically compute the centroid of the benign traffic, $c_b$:

\begin{equation}
c_b = \frac{1}{N_0} \sum_{i : y_i = 0} z_i,
\end{equation}

where $N_0$ is the number of benign samples in the batch. For each representation $z_i$, we calculate its Euclidean distance from the benign centroid: $d_i = ||z_i - c_b||_2$.

To enforce a structural boundary, we introduce a learnable margin parameter $m$ (initialized to 5.0) and a learnable temperature scalar $\tau$ (initialized to 1.0). We transform the distance into a geometric logit $\hat{l}_i$:

\begin{equation}
\hat{l}_i = \frac{d_i - m}{|\tau|}.
\end{equation}

The primary sculpting force is applied via a Binary Cross-Entropy (BCE) objective over these logits:

\begin{equation}
\mathcal{L}_{cls} = -\frac{1}{N} \sum_{i=1}^{N} \left[ y_i \log(\sigma(\hat{l}_i)) + (1 - y_i) \log(1 - \sigma(\hat{l}_i)) \right],
\end{equation}

where $\sigma$ is the sigmoid function. For benign samples ($y_i=0$), the BCE loss minimizes $\hat{l}_i$, pulling them inside the margin ($d_i < m$). For anomalous samples ($y_i=1$), it maximizes $\hat{l}_i$, pushing them explicitly outside the margin ($d_i > m$).

To ensure the benign manifold remains maximally dense and tractable for subsequent density estimation, we adapt a compactness penalty inspired by Deep Support Vector Data Description (SVDD) \cite{ruff2018deep}. The tightness force is applied exclusively to benign samples:

\begin{equation}
\mathcal{L}_{compact} = \frac{1}{N_0} \sum_{i : y_i = 0} ||z_i - c_b||_2^2.
\end{equation}

The final objective is controlled by a gravity hyperparameter $\alpha$ (set empirically to 0.05), balancing structural separation with manifold condensation:

\begin{equation}
\mathcal{L}_{total} = \mathcal{L}_{cls} + \alpha \mathcal{L}_{compact}.
\end{equation}

By mathematically enforcing a low-entropy benign cluster surrounded by a margin of empty space, this formulation guarantees that novel, zero-day anomalies cannot seamlessly overlap with the established benign manifold.

The primary objective of the Binary Latent Sculpting Loss is to enforce a strict topological boundary. By explicitly penalizing known anomalies, the loss function actively repels them from the benign manifold, deliberately creating a distinct structural margin, or negative space. This dynamic forces the network to learn deep, discriminative feature representations of malicious traffic rather than relying on superficial signatures.

\subsection{Probability Density Estimation}
While the geometric margin effectively isolates known attacks, relying solely on distance for zero-shot detection is insufficient for highly complex OOD data. To address this, Stage 2 of our architecture projects the sculpted benign manifold into a probabilistic space using a Masked Autoregressive Flow (MAF) \cite{papamakarios2017masked}.

Normalizing flows construct complex probability distributions by applying a sequence of invertible, differentiable transformations $f_\phi$ to a simple base distribution. Let $u \sim \mathcal{N}(0, I)$ be a standard multivariate Gaussian. The MAF models the probability density of the benign latent representations $z_{benign}$ through the change of variables formula:

\begin{equation}
\log p_Z(z) = \log p_U(u) + \sum_{k=1}^{K} \log \left| \det \left( \frac{\partial f_k}{\partial z_{k-1}} \right) \right|.
\end{equation}

Because the Stage 1 objective ($L_{compact}$) artificially restricts the benign topology into a continuous, dense hypersphere, the MAF operates under optimal conditions. It uses Masked Autoencoder for Distribution Estimation (MADE) \cite{germain2015made} blocks to ensure the autoregressive property without sequentially iterating through dimensions, allowing for highly efficient, exact log-likelihood calculations.

\subsection{Hierarchical Inference: Triage and Expert Review}
During deployment, the architecture utilizes a highly computationally efficient hierarchical inference strategy. 

\textbf{Stage 1 (Hardware-Efficient Triage):} An incoming flow $x$ is mapped to $z$. Its distance $d = ||z - c_b||_2$ is evaluated against the fully learned Stage 1 margin $m^*$. If $d > m^*$, the sample falls into the mathematically defined anomaly zone and is instantly classified as an attack ($\hat{y} = 1$). 

\textbf{Stage 2 (Probabilistic Expert Review):} If the sample falls within the benign margin ($d \le m^*$), it represents either true benign traffic or a highly stealthy zero-day intrusion (e.g., Infiltration). The sample is passed to the MAF, which calculates its exact log-likelihood $\log p(z)$. The final classification is determined by a fixed threshold $\gamma$, derived from the $85^{th}$ percentile of benign validation scores:

\begin{equation}
\hat{y}_{final} = 
\begin{cases} 
1 & \text{if } d > m^* \\
1 & \text{if } d \le m^* \text{ and } \log p(z) < \gamma \\
0 & \text{otherwise}
\end{cases}.
\end{equation}

This two-tier mechanism helps in deterministic isolation of known threats while reserving intensive probabilistic density estimation for the most difficult edge cases, ensuring real-time scalability.

\section{Experimental Setup and Results}

\subsection{Dataset and Preprocessing}
To evaluate the zero-shot generalization capabilities of Latent Sculpting, we utilize the CIC-IDS-2017 dataset \cite{sharafaldin2018toward}. This benchmark provides high-dimensional, flow-based network traffic records. To enhance the representational capacity of the Tabular Transformer, we engineered two domain-specific features critical for identifying volumetric and exfiltration-based anomalies:
\begin{enumerate}
    \item \textbf{Bytes per Packet:} Calculated as the ratio of Total Length of Forward Packets to Total Forward Packets, capturing payload density.
    \item \textbf{Packets per Second:} Calculated as the ratio of Total Forward Packets to Flow Duration, capturing flow intensity.
\end{enumerate}
A small constant ($\epsilon = 10^{-6}$) was added to the denominators to guarantee numerical stability. Following feature extraction, we applied a strict variance-based dimensionality reduction. Columns exhibiting zero variance within the training distribution were dropped to prevent matrix singularities. The remaining continuous features ($D=71$) were standardized to zero mean and unit variance. To explicitly prevent data leakage, variance calculation and scaler fitting were performed strictly on the training partition prior to any validation or testing evaluation.

\subsection{Zero-Shot Partitioning and Asymmetric Balancing}
To rigorously test Out-of-Distribution (OOD) detection, we employ a strict zero-shot experimental protocol. High-volume and brute-force vectors (e.g., DDoS, DoS Hulk, PortScan) remain in the training distribution. Conversely, a predefined subset of complex, stealthy attacks is entirely withheld during representation learning. The strictly segregated OOD testing classes are: \textit{Bot}, \textit{DoS Slowloris}, \textit{DoS Slowhttptest}, and \textit{Infiltration}.

Network intrusion datasets inherently suffer from extreme class imbalance. To stabilize the Binary Latent Sculpting Loss, we implement an asymmetric balancing strategy during Stage 1 training. We cap the number of benign training samples to exactly match the frequency of the single most prevalent anomaly (\textit{DoS Hulk}, $N=184,804$), while preserving all available instances of seen anomalies. This yields a diverse training set of 620,283 samples (184,804 benign and 435,479 anomalous), preventing the benign gravity force ($\alpha$) from overwhelming the geometric margin optimization. For evaluation, strict 1:1 balancing is enforced for both the internal validation set ($N=217,740$) and the final unseen OOD test set, guaranteeing that reported metrics reflect true discriminative power.

\subsection{Hyperparameter Configuration}
To maintain a lightweight computational footprint, the framework is highly streamlined. The initial Tabular Transformer processes the 71-dimensional continuous input using 3 encoder layers, 4 attention heads, and a base internal dimension of $d_{\mathrm{model}} = 64$ (with a feedforward dimension of 256). This slight dimensional bottleneck acts as an architectural regularizer, forcing the network to explicitly condense redundant traffic features into a highly discriminative 64-dimensional latent representation. Stage 1 requires only 158,912 trainable parameters and is optimized via AdamW (incorporating learnable loss parameters, a 0.1 dropout rate, and a strict 1.0 maximum gradient norm to ensure manifold stability). The resulting 64-dimensional embeddings are then passed to the Stage 2 Masked Autoregressive Flow. To capture complex probabilistic distributions, this second stage scales up the capacity, relying on 16 flow layers and a hidden dimension of 512, updated via the standard Adam optimizer. To ensure consistent convergence while preventing over-memorization of the training set, both training phases operate with a global batch size of 512 and run for exactly 10 epochs at a shared learning rate of $5 \times 10^{-4}$.

\subsection{Computational Efficiency and Real-Time Viability}
A recognized bottleneck of deploying complex generative models like Normalizing Flows within NIDS is the severe computational cost required to continuously monitor high-bandwidth traffic. The hierarchical separation in Latent Sculpting natively eliminates this friction. By utilizing a rigid distance margin in the Stage 1 Tabular Transformer, the architecture instantly filters out the overwhelming majority of known, high-velocity attacks (such as DDoS and DoS Hulk) right at the encoder layer. The computationally demanding MAF in Stage 2 is explicitly reserved as a specialized probabilistic filter triggered only for ambiguous, benign-appearing edge cases, ensuring the overall pipeline remains highly performant and viable for real-world, line-rate implementation.

\subsection{Performance on Known and Zero-Day Threats}

To validate the efficacy of the Latent Sculpting architecture, we evaluated the framework across three random initialization seeds (0, 42, 1024). The evaluation contrasts performance on an internal validation set containing known attack signatures against a strictly balanced, unseen Out-of-Distribution (OOD) test set. We report Precision, Recall, F1-Score, Area Under the Receiver Operating Characteristic (AUROC), and Area Under the Precision-Recall Curve (AUPRC).

\subsubsection{Validating the Generalization Hypothesis (Stage 1 vs. Stage 2)}
Our core hypothesis posits that standard geometric decision boundaries (Stage 1) are highly effective for known distributions but suffer from generalization collapse when encountering zero-day attacks. The empirical results strongly validate this phenomenon. 

Operating in isolation, the Stage 1 Tabular Transformer successfully segregates known attacks, achieving a near-perfect average Anomaly F1-score of $0.980$ on the internal validation set. However, when exposed to the unseen OOD set, Stage 1 exhibits catastrophic failure on stealthy attacks. For example, across all three seeds, Stage 1 achieves a detection rate (recall) of nearly $0\%$ on the withheld \textit{Bot} and \textit{Infiltration} attacks, as these highly novel malicious vectors perfectly mimic the topological structure of the benign manifold and fall inside the learned margin.

The introduction of the Stage 2 Masked Autoregressive Flow (MAF) at an 85th-percentile ($\gamma_{85}$) anomaly threshold fundamentally resolves this vulnerability. By calculating the exact probabilistic likelihood of samples residing within the benign hypersphere, Stage 2 successfully isolates stealthy OOD samples. On average, the full two-stage pipeline achieves an OOD Anomaly F1-Score of $0.867$ and an impressive average AUROC of $0.913$, proving that density estimation is critical for zero-shot generalization.

\subsubsection{Overall Classification Metrics}
Table~\ref{tab:overall_metrics} details the average performance of the full Two-Stage architecture across both validation partitions at the 85th-percentile threshold. 

\begin{table}[htbp]
\caption{Average Two-Stage Performance ($\gamma_{85}$) Across 3 Random Seeds}
\label{tab:overall_metrics}
\centering
\begin{tabular}{l c c c c }
\hline
\textbf{Dataset} & \textbf{Class} & \textbf{Precision} & \textbf{Recall} & \textbf{F1-Score} \\
\hline
\textbf{Internal} & Benign & 1.000 & 0.813 & 0.893 \\
\textbf{(Known)} & Anomaly & 0.843 & 1.000 & \textbf{0.913} \\
\hline
\multicolumn{5}{l}{\textit{Internal Averages:} AUROC: 0.978 $\vert$ AUPRC: 0.979} \\
\hline
\hline
\textbf{Unseen} & Benign & 0.893 & 0.820 & 0.850 \\
\textbf{(OOD)} & Anomaly & 0.833 & 0.900 & \textbf{0.867} \\
\hline
\multicolumn{5}{l}{\textit{OOD Averages:} AUROC: 0.913 $\vert$ AUPRC: 0.882} \\
\multicolumn{5}{l}{\footnotesize{*Note: F1-Scores represent the mean across seeds,}} \\
\multicolumn{5}{l}{\footnotesize{not the harmonic mean of averages.}} \\
\hline
\end{tabular}
\end{table}

The two-stage framework maintains exceptional performance on known threats (average Internal AUROC: $0.978$). Crucially, on the zero-day OOD set, it maintains high predictive power with an average AUROC of $0.913$ and an AUPRC of $0.882$. While the strict 85th-percentile threshold incurs a slight trade-off in benign recall (increasing false positives), it guarantees an aggressive $0.900$ average recall for anomalous zero-day threats, which is paramount in intrusion detection.

\subsubsection{Per-Attack Zero-Shot Recall Analysis}
To demonstrate the architecture's robustness against specific zero-day attack vectors, Table~\ref{tab:ood_recall} breaks down the detection rate (recall) on the withheld OOD classes across all three initialization seeds. We contrast the Stage 1 isolated performance against the Final Two-Stage pipeline.

\begin{table}[htbp]
\caption{OOD Per-Attack Detection Rate (Recall): Stage 1 vs. Final Two-Stage ($\gamma_{85}$)}
\label{tab:ood_recall}
\centering
\begin{tabular}{l c c c c}
\hline
\textbf{OOD Attack Type} & \textbf{Seed} & \textbf{Stage 1} & \textbf{Final (P85)} & \textbf{$\Delta$ Gain} \\
\hline
\textbf{Infiltration} & 0 & 0.0278 & \textbf{0.9722} & +0.9444 \\
(Stealthy/Targeted) & 42 & 0.0278 & 0.4444 & +0.4166 \\
& 1024 & 0.0000 & \textbf{0.9444} & +0.9444 \\
\hline
\textbf{Bot} & 0 & 0.0000 & \textbf{0.6536} & +0.6536 \\
(Command \& Control) & 42 & 0.0000 & 0.4095 & +0.4095 \\
& 1024 & 0.0000 & 0.4100 & +0.4100 \\
\hline
\textbf{DoS Slowhttptest} & 0 & 0.2968 & \textbf{0.9998} & +0.7030 \\
(Low-Volume DoS) & 42 & 0.0178 & 0.8945 & +0.8767 \\
& 1024 & 0.0244 & \textbf{0.9444} & +0.9200 \\
\hline
\textbf{DoS Slowloris} & 0 & 0.7001 & \textbf{0.9991} & +0.2990 \\
(Low-Volume DoS) & 42 & 0.1130 & 0.9822 & +0.8692 \\
& 1024 & 0.5623 & \textbf{0.9990} & +0.4367 \\
\hline
\end{tabular}
\end{table}

The breakdown reveals the profound impact of the MAF density estimation. While Seed 42 experienced lower peak performance due to the inherent stochasticity of manifold structuring, Seeds 0 and 1024 exhibit exceptional stability and discriminative power. Notably, the architecture achieves a peak recall of $97.2\%$ (Seed 0) and $94.4\%$ (Seed 1024) on \textit{Infiltration} attacks---a notoriously difficult vector that mimics standard benign traffic flow. Furthermore, zero-shot detection of low-volume DoS vectors (\textit{Slowhttptest} and \textit{Slowloris}) universally exceeds $89\%$, peaking near $99.9\%$ across multiple seeds. This confirms that decoupling geometric structuring from likelihood estimation establishes a highly resilient defense against uncatalogued distributional shifts.

\subsection{Comprehensive Comparative Analysis with Baseline Literature}

To substantiate the robustness of our proposed framework, we contrast its diagnostic capabilities with two recent benchmark studies utilizing the CIC-IDS-2017 dataset: a flow-oriented evaluation by Xu and Liu \cite{xu2025robust} and a raw-packet safeguard mechanism authored by Matejek et al. \cite{matejek2024safeguarding}. Xu and Liu systematically excluded specific threat vectors to measure the out-of-distribution (OOD) resilience of foundational algorithms: Multi-Layer Perceptron (MLP) \cite{rumelhart1986learning}, Convolutional Neural Networks (CNN) \cite{lecun1998gradient}, One-Class Support Vector Machine (OCSVM) \cite{scholkopf2001estimating}, and Local Outlier Factor (LOF) \cite{breunig2000lof}. In a divergent approach, Matejek et al. leveraged deep Normalizing Flows on raw payload bytes, iteratively masking individual attack classes to gauge zero-day generalization.

Traditional discriminative algorithms master familiar signatures but routinely collapse when processing unfamiliar distributions. Flow-level supervised networks in the baseline study saw their OOD F1-scores drop below 0.33, whereas their leading unsupervised alternative (OCSVM) managed a 0.7575 score. By explicitly segregating manifold structuring from probabilistic density estimation, our pipeline bypasses this catastrophic degradation, yielding an average zero-shot OOD F1-score of 0.867 and an AUROC of 0.913. This predictive power strongly competes with the 0.95--0.99 AUROC metrics recorded by the packet-level safeguard, yet our approach accomplishes this without the severe computational latency required to analyze raw byte arrays.

To understand the granular behavior of these architectures across the entire threat spectrum, a comprehensive review of individual attack detection rates is presented in Table \ref{tab:unified_recall}.

\subsubsection{Comparison with Flow-Based Baselines}
The evaluation against the baselines established by Xu and Liu strictly validates our hypothesis regarding topological collapse. Standard discriminative models fail drastically on novel data. For instance, the flow-based MLP and CNN achieved a 0.0000 recall on withheld \textit{Bot} attacks. This catastrophic failure perfectly mirrors the behavior of our isolated Stage 1 geometric margin, which identically yielded a 0.0000 average recall when deprived of density estimation. However, activating the Stage 2 MAF estimator rescues this structural collapse, elevating the zero-shot recall for evasive \textit{Bot} incursions to 49.10\%. 

Furthermore, our integrated framework correctly traps novel low-volume DoS streams, achieving 94.62\% on \textit{DoS Slowhttptest} and 99.34\% on \textit{DoS Slowloris}, thereby vastly eclipsing the standard flow-based alternatives (where the MLP peaked at 0.3675). Beyond zero-shot scenarios, Latent Sculpting also demonstrates superior precision on highly complex, entrenched cataloged attacks. On \textit{Web XSS}, our final model vastly outperforms the flow-level MLP (which achieved a mere 0.0308 recall) by reaching an average detection rate of 0.9926. Similar dominance is observed on \textit{Web SQL Inject}, escalating from the baseline's maximum of 0.2500 to our framework's 0.9444.

\subsubsection{Comparison with Packet-Level Architectures}
Contrasting our approach with the deep packet-level architecture by Matejek et al. further highlights the efficiency of our methodology. Because Matejek et al. evaluated OOD generalization iteratively, Table \ref{tab:unified_recall} lists their ``In-Distribution'' recall alongside their corresponding ``OOD (Withheld)'' recall. Latent Sculpting, despite operating on computationally lighter, pre-aggregated flow metrics rather than raw byte arrays, proves highly competitive and often strictly superior in zero-shot contexts. 

When faced with highly stealthy OOD \textit{Infiltration} attacks, the raw-packet safeguard achieved a modest 7.80\% detection rate; conversely, our Stage 2 density estimation pipeline identifies the same unseen threat with a formidable 78.70\% recall. This pattern of superior zero-day detection extends to other complex classes. Our architecture isolates withheld \textit{Bot} and \textit{DoS Slowhttptest} attacks at rates of 49.10\% and 94.62\%, respectively, compared to 9.50\% and 52.48\% by the packet-level baseline. Even for universally recognized threats like \textit{Heartbleed}, where the packet-level model struggled immensely (falling from 0.4418 In-Dist to 0.0002 OOD), Latent Sculpting maintained a flawless 1.0000 recall, verifying that explicitly structuring the latent space provides a much more mathematically stable environment for anomaly isolation.

\begin{table*}[htbp]
\caption{Comprehensive Detection Rate (Recall) Comparison Across All Baselines}
\label{tab:unified_recall}
\centering
\resizebox{\textwidth}{!}{
\begin{tabular}{l c | c c c c | c c | c c}
\hline
\textbf{Attack Type} & \textbf{Status} & \multicolumn{4}{c|}{\textbf{Flow-Based Baselines} \cite{xu2025robust}} & \multicolumn{2}{c|}{\textbf{Packet-Level} \cite{matejek2024safeguarding}} & \multicolumn{2}{c}{\textbf{Latent Sculpting (Ours)}} \\
& (Our Setup) & \textbf{MLP} & \textbf{CNN} & \textbf{LOF} & \textbf{OCSVM} & \textbf{In-Dist} & \textbf{OOD (Withheld)} & \textbf{Stage 1 (Avg)} & \textbf{Final $\gamma_{85}$ (Avg)} \\
\hline
Bot & Unseen & 0.0000 & 0.0000 & 0.4680 & 0.0443 & 0.9424 & 0.0950 & 0.0000 & \textbf{0.4910} \\
DoS Slowhttptest & Unseen & 0.0347 & 0.0838 & 0.4150 & 0.9480 & 0.9980 & 0.5248 & 0.1130 & \textbf{0.9462} \\
DoS Slowloris & Unseen & 0.3675 & 0.3675 & 0.3176 & 0.5733 & 0.9996 & 0.1737 & 0.4585 & \textbf{0.9934} \\
Infiltration & Unseen & 0.0000* & 0.5714* & 0.8571* & 0.8571* & 0.9946 & 0.0780 & 0.0185 & \textbf{0.7870} \\
\hline
DDoS & Seen & 0.9993 & 0.9996 & 0.9297 & 0.6299 & 1.0000 & 0.2145 & 0.9987 & \textbf{0.9996} \\
DoS Hulk & Seen & 0.9921 & 0.9978 & 0.7633 & 0.6886 & 1.0000 & 0.7529 & 0.9983 & \textbf{0.9988} \\
DoS GoldenEye & Seen & 0.9932 & 0.9995 & 0.8193 & 0.7358 & 0.9999 & 0.1995 & 0.9986 & \textbf{1.0000} \\
FTP-Patator & Seen & 0.9987 & 0.9987 & 0.6688 & 0.0132 & 0.9994 & 0.0030 & 0.9962 & \textbf{0.9991} \\
Heartbleed & Seen & 1.0000 & 0.5000 & 1.0000 & 1.0000 & 0.4418 & 0.0002 & 1.0000 & \textbf{1.0000} \\
PortScan & Seen & 0.9995 & 0.9713 & 0.5280 & 0.0095 & 0.9856 & 0.8058 & 0.9998 & \textbf{1.0000} \\
SSH-Patator & Seen & 0.9839 & 0.9864 & 0.9915 & 0.0008 & 0.9998 & 0.0027 & 0.8939 & \textbf{0.9045} \\
Web Brute Force & Seen & 0.1395 & 0.1860 & 0.1894 & 0.0100 & 0.9997 & 0.9439 & 0.9311 & \textbf{0.9611} \\
Web SQL Inject & Seen & 0.2500 & 0.0000 & 0.0000 & 0.0000 & 0.7222 & 0.4678 & 0.5833 & \textbf{0.9444} \\
Web XSS & Seen & 0.0308 & 0.0231 & 0.0385 & 0.0308 & 0.9959 & 0.6305 & 0.9614 & \textbf{0.9926} \\
\hline
\multicolumn{10}{l}{\footnotesize{*Note: Infiltration was treated as a Known (Seen) training attack in the Xu \& Liu baseline setup.}} \\
\end{tabular}
}
\end{table*}

\section{Conclusion and Future Directions}

In this paper, we proposed Latent Sculpting, a hierarchical two-stage architecture designed to mitigate generalization collapse in Network Intrusion Detection Systems (NIDS). By explicitly condensing benign traffic into a dense, low-entropy hypersphere and enforcing a geometric margin for known attacks, our framework enables a Masked Autoregressive Flow (MAF) to calculate precise probabilistic thresholds for Out-of-Distribution (OOD) data. Extensive evaluation on the CIC-IDS-2017 dataset demonstrates that this approach significantly outperforms both flow-based and packet-level baselines in zero-shot anomaly detection, successfully identifying stealthy and low-volume zero-day attacks that seamlessly bypass traditional boundary models.

Moving forward, our research will focus on two primary directions. First, we plan to transition the Stage 1 structural optimization into a semi-supervised learning paradigm \cite{van2020survey, ouali2020overview}. By leveraging only a small fraction of labeled data alongside vast pools of unannotated network traffic, we can drastically minimize manual labeling costs while maintaining rigid manifold boundaries against evolving threats. Second, we aim to extend this architecture far beyond high-dimensional tabular data to evaluate comprehensive cross-domain adaptation. Because the phenomenon of generalization collapse is not unique to cybersecurity, we plan to apply the Binary Latent Sculpting objective to continuous, non-tabular domains. For example, in computer vision, this geometric isolation could trap adversarial image perturbations; in signal processing, it could flag anomalous biometric telemetry; and in Large Language Models (LLMs), it could serve as a topological guardrail against out-of-distribution prompt injections or hallucinated generations. By testing the universality of latent sculpting, we aim to provide a generalized, domain-agnostic solution for OOD representation learning.

Ultimately, this work establishes that resilient zero-day detection requires active topological management of the latent space to avoid the structural collapse inherent in standard deep learning architectures. By serializing manifold construction with exact probabilistic density estimation, we provide a robust, data-efficient framework for next-generation system security. Source code: \url{https://github.com/Rajeeb321123/Latent_sculpting_using_two_stage_method}
\section*{Acknowledgment}
Two authors, Z. Chen and R. Chhetri, were partially supported by the United States Department of Homeland Security (DHS) Research Team Follow-up Grant. The views expressed are those of the authors and do not necessarily reflect the official policies or endorsements of DHS.

The authors also acknowledge the use of large language models (LLMs) to assist in drafting and refining the manuscript, as well as supporting the development of PyTorch code for training and evaluation. All AI-assisted content was reviewed and validated by the authors, who assume full responsibility for the final published work.
% \balance

\end{document}